# Phonemic Representation and Transcription for Speech to Text Applications for Under-resourced Indigenous African Languages: The Case of Kiswahili

Awino, Ebbie[2]; Wanzare, Lilian[1,3]; Muchemi, Lawrence[2]; Wanjawa, Barack[2]; Ombui, Edward[4]; Indede, Florence[3]; McOnyango, Owen[3]; Okal, Benard[3]

1 – corresponding author, 2 – affiliation University of Nairobi, Kenya, 3 – affiliation Maseno University, Kenya. 4 – affiliation Africa Nazarene University, Kenya

**Abstract**

Building automatic speech recognition (ASR) systems is a challenging task, especially for under-resourced languages that need to construct corpora nearly from scratch and lack sufficient training data. It has emerged that several African indigenous languages, including Kiswahili, are technologically under-resourced. ASR systems are crucial, particularly for the hearing-impaired persons who can benefit from having transcripts in their native languages. However, the absence of transcribed speech datasets has complicated efforts to develop ASR models for these indigenous languages. This paper explores the transcription process and the development of a Kiswahili speech corpus, which includes both read-out texts and spontaneous speech data from native Kiswahili speakers. The study also discusses the vowels and consonants in Kiswahili and provides an updated Kiswahili phoneme dictionary for the ASR model that was created using the CMU Sphinx speech recognition toolbox, an open-source speech recognition toolkit. The ASR model was trained using an extended phonetic set that yielded a WER and SER of 18.87% and 49.5%, respectively, an improved performance than previous similar research for under-resourced languages.

## 1       Introduction

There is a common desire to maintain, protect, sustain, and revitalize native languages in many regions of the world where they are spoken.  Language is a means to transform and invigorate culture. On the other hand, technology has an impact on how individuals and groups communicate, acquire knowledge, and think. It benefits society and impacts how people communicate with one another every day. The development of language technologies may include process improvement, technology invention, innovation, or diffusion. Basic rule-based models of language technology have been replaced over time by more resource-intensive techniques like machine learning and deep learning. The requirement for resources such as text and speech corpora, phonemic-dictionaries and speech transcription datasets is critical for any viable voiced-based language-technology computer system. A requisite for developing such systems is adequate data to design and train models. Most indigenous African languages are categorized as resource scarce, meaning there is no adequate training data to build models that support African languages. Therefore, the indigenous languages are underrepresented in Artificial Intelligence with an impact of the speakers suffering from poor digital inclusivity, lack of participatory decision making, inequity and poor information access.

To enhance the capacity of these languages and more so in the development of speech-to-text and text-to-speech systems, there needs to be a better understanding of the requirements for these tasks especially from a linguistic as well as a computing perspective. Methodologies have to be explored from best practices and harmonized to enable faster and accelerated development. This paper reports part of research activities carried out under Kencorpus Project (Wanjawa et al., 2022).  From a linguistic perspective, we investigated the phonemic representation of Kiswahili, an indigenous

African language that is widely spoken in Eastern Africa and parts of Central and Southern Africa. It is appreciated that pronunciation varies from one region to another and also varies depending on the context. The key research question therefore is; what phonemic set should be used to obtain the most optimum performance in speech to text systems? As a tangible output of this we sought to create a comprehensive phonemic set to obtain better performances in Speech to text Systems. From a computational perspective, we sought to investigate the best practices for collecting and curating speech data to create a speech corpus, creation of a comprehensive words-phonemes dictionary for Kiswahili and the creation of transcription files. Further an investigation was carried out to bring forth the most suitable open-source speech to text libraries for processing such data, and also find optimum quantity and varieties necessary for development of a viable speech-to-text system.

The rest of the paper is organized as follows: in Section 2, we discuss background work on transcription, Kiswahili language and Kiswahili phonemes, and related work on ASR systems. Then in Section 3, we discuss the methodology that includes the model adopted in building the Kiswahili ASR system. Section 4 presents the experiments that were conducted in the study when developing the Kiswahili ASR. Section 5 presents the results of the various models and discussions. Finally Section 6 concludes the paper.

## 2. Background
### 2.1 Concept of transcription

Transcription, in linguistics, is the conversion or transfer of audio files into written formats (Dresing et al., 2015; Bailey, 2008 as cited in Rincon, 2018; International Rescue Committee-IRC, 2018). This conversion is normally done to ease communication. Transcription also helps in converting audio files into written formats in order to be studied and analyzed (Rincon, 2018; McMullin, 2021). As a special skill, transcription helps learners to become very successful writers thus improving on their writing and reading abilities (Deem et al., 1988). Audio files that are normally transcribed during the transcription process include words heard by the transcriber. Transcription is also an integral stage in the creation of speech corpus normally used in speech-to-text systems.

Transcription process is very cumbersome in that a transcriber may spend between 3-8 hours just transcribing one audio file (McMullin, 2021). Therefore, transcription tends to be complicated and time consuming (Spencer et al., 2006). Transcription requires an oscillation between a realistic and possible representation of the audio files, thus any transcript will not fully represent the real interview situation (Dresing et al., 2015). That means some deviations and errors will normally occur. The transcriber therefore needs to look for various ways of minimizing these errors (Spencer et al., 2006). Common samples of audio files that are normally transcribed include conversations, interviews and dictations among others (Dresing et al., 2015). Though there are technologies that translate speech into written texts especially in the highly resourced languages such as English, as can be witnessed in YouTube or even smartphones, these technological resources do not effectively cover other low-resource languages in Africa such as Kiswahili among other indigenous languages (Gelas et al., 2012).

### 2.2 Transcription Typologies

In order to develop effective transcriptions for use in speech to text systems for our selected case-study Kiswahili language, there is a need for deeper understanding of the typologies and procedures for transcription. Transcription may be of the following three types: verbal/verbatim transcription, edited transcription, and intelligent transcription (Speechpad team 2013 as cited in Rincon, 2018). Verbatim means that the text is transcribed as heard by including all sounds/words spoken, laughs, background noise, jargons, broken or incomplete phrases and words included. In this case, the transcript is purely a replica of the audio recording. In the second type, edited transcription, some parts of the audio files are omitted while the intended meaning is not altered. Such type of transcription tends to exclude speech errors, false starts and other sounds such as interjections among others. The transcriber picks only what is felt to be important. This type of transcription is very prevalent in speeches, conferences and seminars among others. The final type of transcription is intelligent transcription. This is a special type where the transcriber tends to add personal pronouns and other omitted articles especially in the speeches that were done in a rush. The grammar and other jargon are always written in standard form. This kind of transcription is common with academic works. We selected the verbatim transcription approach.

### 2.3 Transcription Procedures

Transcription can be done by an individual or many individuals on one audio file or various files. Therefore, we may have single or parallel transcriptions (Spencer et al., 2006) that are done for purposes of analysis. In single transcription, one person transcribes an audio file alone and then checks against the original as many times as possible. On the other hand, parallel transcription requires that one audio file is transcribed by several transcribers. Correction of places with disagreements is done thereafter. Finally, additional checks of the entire transcript are done against the original audio file. In our case we selected a rapid method where we had many trained personnel singularly transcribe audios or texts and a few others check for quality on all the transcriptions.

Recording of audio files is first done before the transcribing process can commence. There are best practices for recording files in readiness for transcription, which provide for the following considerations (Moore et al., 2017); first is what and why recording is being done and secondly a decision on whether to record either as audio or video. Finally, the type of recording equipment for use that is appropriate for audio or video is considered e.g. smartphones, tablets, video or audio recorders. There is usually a need for use of good microphones that are clearly linked to a video camera or audio recorder when recording in a very noisy environment.

The transcription process can then follow after obtaining the audio recording. This is generally a five-step process (Moore et al., 2017). First, you have to organize data into a corpus. This can be done by using Computer Assisted Qualitative Data Analysis Software (CAQDAS) packages such as Nvivo or Atlas. This results in appropriate naming conventions for the data files. The second step is to synchronize audio files to decide on which data file to use, then making a decision on what to transcribe. It is always recommended that you do not transcribe the entire corpus. The fourth step requires the decision on the program to use for transcription e.g. CLAN, ELAN, Nvivo or Atlas. Finally, you create an initial transcription file in which details like prosody, gestures, pauses among others are not included. However, it is also possible to do a fine transcription that includes symbols to represent multimodal features like intonation, stress, rhythm, gestures, gaze among others in the text form. In our case, the transcribers listened to audio clips using a media player, and typed out the text in a text editor. We adopted this approach as most transcribers were unfamiliar with software tools and was faster to use what is already known.

### 2.4 Kiswahili language

Languages in Africa just like other languages world over can be classified in terms of geographical, structural, functional and familial expositions (Mgullu, 1999; King'ei, 2010). Kiswahili, also referred to as Swahili, tends to fall in all these classification categories. Geographically, Kiswahili is an East African language standardized from the Ki-Unguja dialect. Structurally, it is an agglutinating language that accepts affixations. Functionally, the language is both an official and national language of Kenya and Tanzania, and is widely spoken in Democratic Republic of Congo, Mozambique, Somalia, Uganda, Rwanda and Burundi (Gelas et al., 2012). Kiswahili is an official language of communication in the East African Community and African Union summit functions. Kiswahili belongs to the Niger-Congo language family which is spoken by people who inhabit parts of western, central, eastern and southern Africa. Besides Kiswahili, the Niger-Congo family also includes: Yoruba, Igbo, Congo, Luganda,

Kinyarwanda, Makua, Isi-Zulu, Isi-Xhosa, Ndebele, Manding, Wolof, Kikuyu, Kamba and Nyamwezi among others (Webb et al.,2000; Singler, 2011).

Kiswahili is also classified as a Bantu language spoken by about 100M people in the eastern and southern Africa with about 5M native speakers (Akidah, 2013; Gelas et al., 2012). Other scholars give an estimate of about 150M speakers in the East African coast (Gangji et al., 2015). Kiswahili is considered as one of the indigenous languages in Kenya and Tanzania and also spoken in parts of Southern Africa. It is also a taught subject in schools in Kenya and Tanzania, and now Uganda and Rwanda thus playing a key role in teaching in the education sector. Because of its international recognition, the language is currently a very important language used for translation and interpretation purposes.

## 2.5   Kiswahili Phonemes

A phoneme is a fundamental unit of phonological structure, an abstract segment which is one of a set of such segments in the phonological system of a particular language or speech variety. It is the smallest unit which can make a difference in meaning, distinguishing one word from another (Verma et al., 1989; Habwe et al., 2004; Massamba 2012; Mgullu 1999; Trask, 1996). A phoneme is the smallest abstract unit of sound of a specific language. For example, in Kiswahili, /t/ in *taka* and the initial 'k' in *kaka* are phonemes, since they make a difference in meaning in Kiswahili. This smallest unit is not further divisible however it can act singly or in combination to form syllables, words and then sentences.

Phonemes in any given language can appear as either vowels or consonants. The consonants vary both in number and examples. However, there are similar or uniform phonemes across languages in the world. Kiswahili language also has both the vowel and consonant phonemes and is one of the most documented African languages (wa Mberia, 2015). Kiswahili scholars tend to vary in their opinions regarding the number and examples of Kiswahili phonemes, especially the consonants.

There are 5 Kiswahili vowels namely: /a, e, i, o, u/ (Ashton, 1951; Whiteley, 1956; Polome, 1967; Nchimbi, 1995; Mgullu, 1999; Habwe et al., 2004; Kihore et al., 2009; Obuchi et al., 2010) though other scholars indicate the presence of 5 vowel phonemes and 2 semi vowels /w, y/ (Akidah, 2013). These Kiswahili vowel phonemes are phonetically presented as /i, ɛ, a, ɔ, u/. However, variations tend to emerge in the number of consonants together with their examples in Kiswahili. Earlier linguists mentioned between 32-37 existing Kiswahili phonemes, both vowel and consonants. Some of the listing and number of phonemes in Kiswahili according to various researchers is shown on Table 2.1 below:

**Table 2.1** – Various Kiswahili consonant phonemes

| Research | Consonants | Total |
|---|---|---|
| Ashton (1951) | /p, b, m, w, mb, f, v, mv, th, dh, j, t, d, s, z, n, r, l, nd, nz, ch, sh, nj, ny, y, k, g, kh, gh, ng', ng, h/ | 32 |
| Whiteley (1956) | /p, ṗ, b, m, w, mp, mb, f, v, t, ṫ, d, n, r, l, s, z, nt, nd, ch, cḣ, j, sh, nj, ny, y, k, k̇, g, ng', nk, ng, h/ | 33 |
| Polome (1967) | /p, ph, b, t, th, d, c, ch, j (ɟ), k, kh, g, f, v, th (θ), dh (ð), s, z, š, gh (ɣ), h, m, n, ny (ɲ), ng' (ŋ), l, r, w, y (j)/ | 29 |
| Nchimbi (1995) | /p, t, ch (tʃ), k, b, d, j (ɟ), g, m, ny (ɲ), ng' (ŋ), n, f, th (θ), s, sh (ʃ), h, v, dh (ð), z, r, l, y (j), w, gh (ɣ), mb, nd, ng, nj/ | 29 |
| Mgullu (1999) | /b, ch (tʃ), d, dh (ð), f, g, gh (ɣ), h, k, l, m, n, ny (ɲ), ng' (ŋ), p, r, s, sh (ʃ), t, th (θ), j (ɟ), v, z, x (kh)/ | 24 |
| Habwe et al. (2004) | /p, b, m, w, f, v, th (θ), dh (ð), t, d, l, r, n, s, z, j (ɟ), ny (ɲ), y (j), sh (ʃ), ch (tʃ), k, g, ng' (ŋ), gh (ɣ), h/ | 25 |

| Kihore et al. (2009) | /b, ch, d, dh (ð), f, g, gh (ɣ), h, j, k, l, m, n, ny (ɲ), ng' (ŋ), p, r, s, sh (ʃ), t, th (θ), v, w, y, z/ | 25 |
|---|---|---|
| Obuchi et al. (2010) | /p, m, f, th (θ), t, l, n, z, ny (ɲ), sh (ʃ), k, ng' (ŋ), h, b, w, v, dh (ð), d, r, s, j, y, ch (tʃ), g, gh (ɣ)/ | 25 |
| Akidah (2013) | /b, b, m, f, v, θ, ð, t, d, n, ɲ, s, z, l, r, j, ʃ, k, g, tʃ, ŋ, ɣ, h/ | 23 |

Though it appears that there are many variations in the number and examples of Kiswahili consonants among Kiswahili scholars, the more agreeable number is 29 (Nchimbi,1995) as shown on Table 2.1. This number and confirmation by examples is supported by more scholars (Mwita, 2007; Choge, 2009; Okal, 2015). These 29 Kiswahili consonant phonemes can be detected in Kiswahili words as provided in **Appendix 1**.

It is worthwhile to note that studies on Kiswahili consonant phonemes are still ongoing (Choge, 2009). In this respect, inclusion of other adapted words from other foreign languages like the Arabic language tends to further increase the number and examples of these phonemes. For instance, the word 'kheri' adapted from Arabic language gives us the phoneme /kh/. Thus, we may also have this phoneme in the Kiswahili phoneme catalog:

/kh/ = **kh**eri

It is important also to remember that Kiswahili consonant phonemes can be represented with one alphabet or compounded. This is referred to as single letter or diagraph presentation (combination of more than one letter but regarded as one phoneme) (Gelas et al., 2012). In many cases Kiswahili phonemes, just like other phonemes among many languages world over, are normally represented with one symbol or alphabet as follows: /a, e, i, o, u/ for vowels and /p, t, k, b, d, j, g, m, n, f, s, h, v, z, r, l, y, w/ for consonants. However, there are also consonant phonemes that are combinatory in nature. For example, /th, dh, gh, kh/ are Kiswahili phonemes that are derived from Arabic language (Treece, 1990; Baldi, 2012). Thus, consonant phonemes that appear to be compounded structurally or that are technically referred to as digraphs are normally regarded as one single phoneme if some two conditions are met. These conditions (Trubetzkoy, 1969) are firstly, only those combinations of sounds whose constituent parts in a language are not distributed over two syllables, are to be regarded as the realization of single phonemes. Secondly, a combination of sounds can be interpreted as the realization of a single phoneme only if it is produced by a homogenous articulatory movement or by the progressive dissolution of an articulatory complex.

Though expositions of earlier studies regarding Kiswahili phoneme representation in the production of dictionary for NLP have been done, there are similarities and deviations from our research. According to Gelas et al., 2012, there are 36 Kiswahili phones namely: /BB, CC, DD, GG, JJ, LL, NN, RR, SS, TT, VV, XX, ZZ, a, b, d, e, f, g, h, i, j, k, l, m, n, o, p, r, s, t, u, v, w y, z/[12]. These phonemes can be presented by words in Kiswahili as per the sample list shown in **Appendix 2.**

## 2.6 Automatic Speech Recognition (ASR) systems

In the last decades, several works have been conducted in order to aid the adoption of various natural language technologies within under-resourced languages. Getao et al. (2006) developed a sentence-based system to transcribe Kiswahili speech to text. The study included four distinct phases including the pre-processing level, the emission model, the lexicon model and the transcriber. At the pre-processing phase, the scholars used the Mel-frequency Cepstral Coefficients (MFCC) to extract features from the speech signals. The emission model that identifies correspondences between the input speech features and a given phone was created using a multi-layer perceptron (MLP) neural network and trained using the back propagation algorithm. The model used spoken sentences as training data that included the acoustic signals generated when the sentences were read and a phonetic transcription of each segment within the acoustic signal.

---

[1] https://raw.githubusercontent.com/getalp/ALFFA_PUBLIC/master/ASR/SWAHILI/lang/lexicon.txt
[2] https://github.com/getalp/ALFFA_PUBLIC/blob/master/ASR/SWAHILI/lang/nonsilence_phones.txt

Additionally, the lexicon model contained the transcribed words, bigram probabilities, and the probability of each transcribed word coming at the beginning of a sentence. Lastly, the transcriber was created using the Viterbi algorithm to facilitate a recognizing function of the words within any particular spoken sentence. However, some of the evidenced challenges of the model developed by Getao et al. (2006) included its understanding of speech from only one specific speaker. The research achieved a 43% word error rate (WER) that was significantly high. The transcription speed and the training rate of the developed model was evidently low. The slow transcription speed resulted in the lexicon model containing limited vocabulary that may have also contributed to the high word error rate.

In another study, Nimaan et al. (2006) developed an automatic transcription model for the Somali language whose data was first normalized due to multi-spelling variants for certain Somali words. The study's acoustic model was developed using the LIA acoustic modeling toolkit. The researchers also developed a language model using Somali trigrams trained using the Somali textual corpus obtained from the CMU and LIA toolkits. The model contained 726,000 bigrams and 1.75 million trigrams of words in the Somali vocabulary and 189,000 bigrams and 996,000 trigrams of root forms in the Somali vocabulary. Nimaan et al. (2006) utilized the LIA large vocabulary speech recognition system (Speeral) for speech decoding and obtained a 32% WER. However, after a spelling normalization the model obtained a 20.9% WER. Additionally, at the root level, the model obtained a 14.2% word-root error rate.

Similarly, Biswas et al. (2019) in their study first had their audio files divided into fixed length files before being used as inputs into the recognizer and semi-supervised retraining. Based on research conducted by Yılmaz et al. (2018b) that established the significance of incorporating artificially generated text using long short-term memory (LSTM) in lowering the perplexity of language models, Biswas et al. (2019) also utilized a trained LSTM network on their Somali acoustic training transcriptions that realized an 11-million-word corpus. Additionally, the researchers utilized semi-supervised training owing to past indications of its ability to improve performance of ASR models designed for under-resourced languages. Biswas et al. (2019) also developed a language model using the SRILM toolkit. The acoustic model, on the other hand, was created using the Kaldi speech recognition toolkit. The researchers established that while time-delay neural network (TDNN) used only half of the model's parameters, it outperformed the CNN-TDNN-BLSTM with WERs of 50.95% and 53.75% respectively. Additionally, that research also found that due to the availability of a large vocabulary set within the Somali language the incorporation of a semi-supervised training data improves performance of the ASR model.

African nations are largely multilingual resulting in significant code-switching among speakers. Yilmaz et al. (2018a) developed a code-switching ASR for five languages out of the possible eleven South African languages. The language model utilized both monolingual and code-switched text resources that were then interpolated with an English monolingual language model to decrease transcription perplexities. The study used various neural network algorithms including TDNN and LSTM to develop the acoustic model. Since most South African languages are tonal, the extraction of pitch data as part of the acoustic features proved vital. That research found that the consideration of pitch features reduced the WER by 1.7%, while adoption of a 5-gram language model decreased the WER by 2.2%.

For some West African languages, Gauthier et al., (2016b) used the Kaldi speech recognition toolkit to develop two ASR systems for Hausa and Wolof, with vowel length contrasts. The study created two acoustic models using two distinct techniques, deep neural networks (DNN) and the classical Hidden Markov Model and Gaussian Mixture Model (CD-HMM/GMM). The language model, on the other hand, used trigrams. The CD-DNN-HMM acoustic model with no incorporation of the vowel length details achieved WERs of 8.0% and 27.2% for Hausa and Wolof respectively. This score was significantly lower than the acoustic models developed using the CD-HMM-GMM that achieved 13.0% and 31.7% WERs for Hausa and Wolof respectively. The researchers established that the high WERs evidenced on Wolof was due to the high perplexity within the Wolof language model due to the presence of only two speakers in the test set in contrast to 10 speakers in the Hausa language model. The differences were also attributed to the presence of more unknown words in the Wolof language model than in the Hausa language model i.e. 5.4% and 0.19% respectively. Additionally, the Wolof ASR lacked normalization of written words that affected the WER such as was evidenced through the low character error rate (CER) that is less sensitive to spelling issues. That research evidenced slight improvement in the Hausa ASR upon the introduction of vowel length contrasts in the acoustic models. The new Hausa acoustic model achieved a WER of 7.9% using the CD-DNN-HMM acoustic model and 12.9% using the CD-HMM-GMM. However, there was a slight degradation of performance of the new Wolof acoustic model.

In Eastern Africa, Mukiibi et al., 2022 developed an ASR system for Luganda using the Coqui Speech-to-Text (STT) model[3] whose core is a recurrent neural network (RNN). They built a language model using the Kenlm toolkit. Their system was based on Radio data and Mozilla common voice datasets for Luganda. Their best performing model got a WER of 33% on common voice data and 47% on radio data.

Closest to this study is work done by Gelas et al., 2012 who developed an ASR model for Kiswahili using the Sphinx Train toolkit[4] that is based on the Hidden Markov Model (HMM) acoustic model with 37 predefined units. They trained a Kiswahili language model using the SRI[5] language model toolkit. Their best performing system uses sub-word units for language modeling, getting a WER of 34.8% in all environments, and WER of 25.9% in good acoustic environment. We follow a similar methodology and improve on the Kiswahili phonemes dictionary and the language model.

## 3     Methodology
### 3.1    Data

This research is based on data which was collected by the KenCorpus project (Wanjawa et al., 2022). The KenCorpus project collected Kiswahili text and speech data, among other languages, from across Kenya. The speech data collected in the project included interviews, conversations, stories, and news among others. For Kiswahili, the KenCorpus project collected 2,585 Kiswahili text files and about 19 hrs of Kiswahili speech. We built a transcription corpus by utilizing both the collected text files and the speech files. Using the collected files, native speakers read out sentences to get text-voice pairs. Similarly, transcribers listened to the collected speech data and typed out what they heard. In the end, our corpus included both read and spontaneous speech corpora.

The read speech corpus consisted of 150 Kiswahili documents and 324 code-switched tweets, some of which were translated to Kiswahili. These texts were first cleaned (converted to lowercase, removal of punctuation marks and split into shorter sentences). The shorter sentences were processed to enable us to get approximately 20-30 seconds of recording when they were read out. The shorter sound clips were chosen for convenience of processing.

---

[3] https://stt.readthedocs.io/en/latest/Architecture.html

[4] cmusphinx.sourceforge.net/

[5] www.speech.sri.com/projects/srilm/

The research recorded 5,652 read out texts (sentences) that represented 26hrs 32min 37sec of speech data. For the spontaneous speech, our corpus contains 119 utterances collected from native Kiswahili speakers in various community settings that represent 59min 13sec of speech data. The total speech dataset of the project therefore was 27hrs 31min 50sec of speech data from 26 speakers (19 female and 7 male). Table 3.1 summarizes the speech data that was built.

**Table 3.1**: Speech Corpus statistics

|  | Sentences | Duration |
|---|---|---|
| Read sentences | 5,652 | 26 hrs 32 mins 37 secs |
| Spontaneous speech | 119 | 59 mins 13 secs |
| **Total** |  | 27hrs 31 mins 50 secs |

### 3.2 Choices made for Kiswahili Transcripts

Though transcription methodology can be single or parallel (Spencer et al., 2006), we adopted the single transcription method in which many Kiswahili transcribers were provided with audio files to transcribe. Each person transcribed an audio file on their own, then rechecked against the original source for as many times as was possible to get the best transcription, as also done by other similar research (Dresing et al., 2015). Our choice of single transcription was convenient for the corpus size and the limited number of transcribers.

Our research adopted the verbatim type of transcription to help in ensuring that Kiswahili words are enabled in the datasets. This is because Kiswahili, just like other African indigenous languages, is still technologically under-resourced. Therefore, all aspects of parts of speech in Kiswahili need to be covered for the ASR system.

### 3.3 Augmentation of the Kiswahili dictionary

Though the 36 Kiswahili phonemes[6] were applicable to our research (Gelas et al., 2012), we had modifications when doing the representations of the Kiswahili phonemes in the dictionary. The first form of dictionary used was the basic form consisting of separation of the consonants and vowels with single spaces. For example, the word *kaka* would have a dictionary entry as *k a k a*. This consonant separation exercise yielded 34 phonemes (29 consonants and 5 vowels) as supported by scholars and provided in **Appendix 1** (Nchimbi,1995; Mwita, 2007; Choge, 2009; Okal, 2015). The basic form dictionary was a random baseline that proved deficient in correctly identifying phonemes in certain words. For example, the word *ndoo* with entry *nd o o* indicates that the double 'o' sounds are pronounced as two separate 'o' sounds.

The second form of baseline dictionary used was a replica of the dictionary created by the African Languages in the Field: speech Fundamentals and Automation (ALFFA) project (Gelas et al., 2012). Unlike the basic form dictionary, the ALFFA dictionary yielded 36 phonemes. In comparison to the basic dictionary the ALFFA dictionary introduced three new phonemes /*nz, kh, mv*/ but omitted the /*ny*/. Similar to the basic dictionary, the ALFFA dictionary, our second baseline with reference to previous work, also proved deficient in correctly identifying phonemes in certain words such as those with the /*ny*/ consonant and double words. For example, the word *abbey* with entry *a b b e y* indicates that the double 'b' consonant is pronounced as two separate 'b' consonants.

---

[6] https://github.com/getalp/ALFFA_PUBLIC/blob/master/ASR/SWAHILI/lang/nonsilence_phones.txt
   https://raw.githubusercontent.com/getalp/ALFFA_PUBLIC/master/ASR/SWAHILI/lang/lexicon.txt

Due to such limitations in the capturing of the entire Kiswahili phonemes, we built an augmented dictionary using the CMU Sphinx English dictionary as reference. English is a high-resourced language with extensive research that has ensured its NLP resources provide higher accuracies in language processing tasks. However, English and Kiswahili vary in how their consonants are sounded. This resulted in some of the English phonemes being maintained in the Kiswahili dictionary, while others such as the consonant /x/ were removed. Similarly, Kiswahili consonants lacking in the English language such as /ny/ were introduced. We observed that significant language comprehension was needed to properly annotate the various Kiswahili words with their proper phoneme sequence to match their pronunciations within the audio files. We applied different representations of the Kiswahili phonemes in the dictionary. This reference exercise yielded 40 phonemes that were utilized in the phonelist file (see Table 3.2 for the full phoneme list). The modifications on the dictionary representation experimentally resulted in better performance for the ASR system.

**Table 3.2**: The 40 Phonemes used for the Kiswahili transcription dictionary

| Phoneme | Sample word | Dictionary for NLP |
|---|---|---|
| AH | au | **AH** UH |
| AA | maana | M **AA** N AH |
| B | bana | **B** AH N AH |
| CH | chakula | **CH** AH K UH L AH |
| D | dada | **D** AH **D** AH |
| DH | dhana | **DH** AH N AH |
| EH | lete | L **EH** T EH |
| EE | mletee | M L EH T **EE** |
| F | funga | **F** UH NG AH |
| G | gonga | **G** OH NG AH |
| GH | ghali | **GH** AH L IH |
| HH | hali | **HH** AH L IH |
| IH | imba | **IH** MB AH |
| II | miiko | M **II** K OH |
| JH | jana | **JH** AH N AH |
| K | kana | **K** AH N AH |
| KH | kheri | **KH** EH R IH |
| L | lala | **L** AH L AH |
| M | maneno | **M** AH N EH N OH |
| MB | mboga | **MB** OH G AH |
| N | nina | **N** IH N AH |
| NG | ngoma | **NG** OH M AH |
| NG' | ng'oa | **NG'** OH AH |
| ND | ndizi | **ND** IH ZIH |
| NY | nyayo | **NY** AH Y OH |
| NJ | njia | **NJ** IH AH |
| OH | ona | **OH** N AH |
| OO | njooni | NJ **OO** N IH |
| P | paka | **P** AH K AH |
| R | rarua | **R** AH **R** UH AH |
| S | sana | **S** AH N AH |
| SH | shika | **SH** IH K AH |
| T | tena | **T** EH N AH |
| TH | thamani | **TH** AH M AH N IH |
| UH | ua | **UH** AH |

| UU | muungwana | M **UU** NG W AH N AH |
| V | vuna | **V** UH N AH |
| W | weka | **W** EH K AH |
| Y | yai | **Y** AH IH |
| Z | zeze | **Z** EH **Z** EH |

### 3.4 Automatic Speech Recognition (ASR) Model for Kiswahili Language

In our research we used CMU Sphinx speech recognition toolkit (CMUSphinx, 2022) to build the Kiswahili language ASR system. The toolkit constitutes a speech decoder and acoustic model trainer that exploit packages including Sphinxtrain, Sphinx, and Pocketsphinx. Sphinxtrain handles the acoustic model training for languages that is primarily based on the Hidden Markov Model (HMM) and the Gaussian Mixture Model (GMM). The developed acoustic model is then utilized either by Sphinx or Pocketsphinx to convert speech signals into readable texts. Pocketsphinx is a smaller version of Sphinx designed for low-resource projects, for instance, embedded systems.

CMU Sphinx is used for two distinct phases – the model development phase and the speech recognition phase. The architecture of CMU Sphinx that we used in this research is depicted on Fig. 3.1 below.

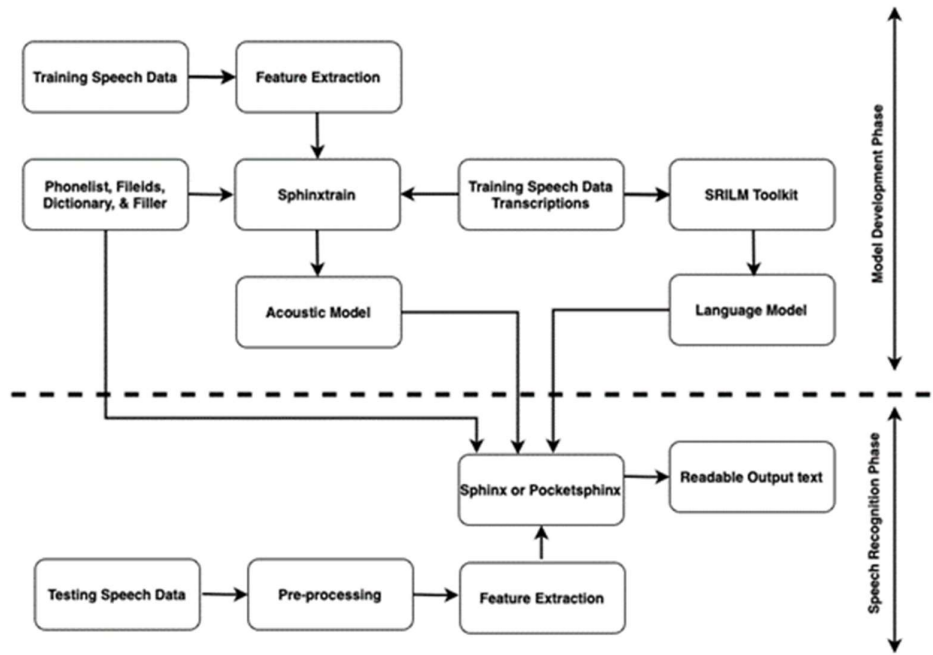

Figure 3.1 – CMU Sphinx speech recognition toolkit architecture (source: Reddy et al., 2015)

CMU Sphinx-4 was also used in the evaluation of the developed acoustic model, language model, and the dictionary. Unlike other decoders such as Pocketsphinx, Sphinx-4 is java-based with a high degree of modularity. Sphinx-4 consists of three distinct sections, being the frontend, the knowledgebase and the decoder. This is depicted in Fig. 3.2 below:

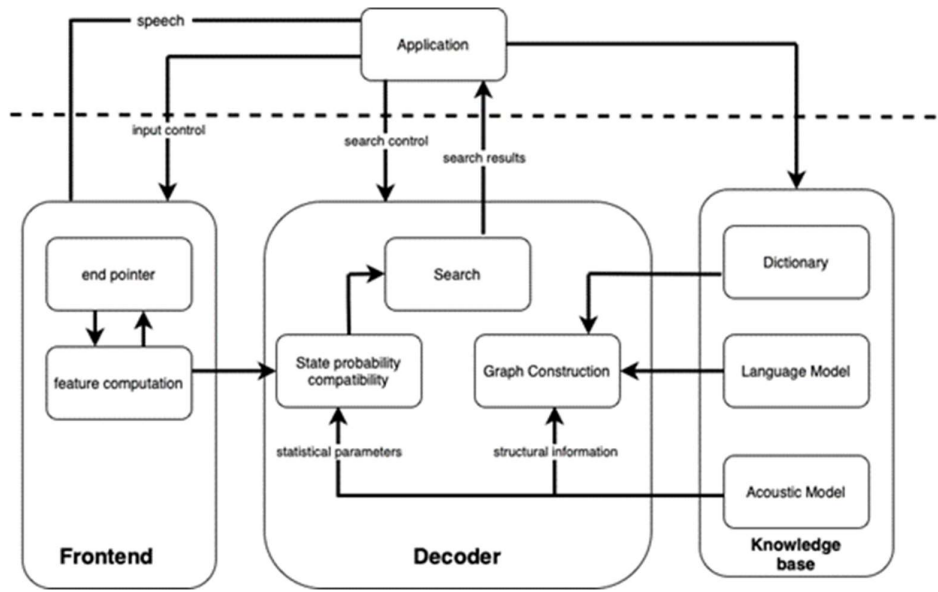

Figure 3.2 – Architecture of the Sphinx-4 system architecture (source: Pantazoglou et al., 2018)

The frontend block includes the endpointer and the feature extractor that takes in speech and extracts significant features from the audio signals. The knowledge base (dictionary, language model, and acoustic model) is used in the creation of a language HMM applied by the decoder block.

The decoder, which performs the actual recognition, is the most significant block of the model. It includes the graph reconstruction module that constructs the system's language HMM. It accomplishes this by first converting the knowledge base's language model to an internal Sphinx-4 understandable format. This new language model in conjunction with the dictionary data and structural data from the acoustic model are then manipulated to develop the language HMM.
The graph construction module's output (language HMM) serves as part of the input to the search module that is then used to determine the pattern to be searched for within the speech data. This search is done through token passing and the results occur at various levels e.g. at the phone or word levels. This search is also highly dependent on the state output probability values computed for each feature vector input in the state probability computation module.

The application module as used in this research was developed in Java. It accepts real-time audio data collection, use of pre-uploaded audio existing in the system or uploading of audio files from the user's device. In either of these three options, the audio files are first pre-processed to conform to the CMU Sphinx required audio formats as detailed in Table 3.3 below.

**Table 3.3**: Audio file characteristics for use with CMU Sphinx

| Audio parameter | Value |
|---|---|
| Format | Wav |
| Channels | Mono |
| Bitrate | 16its |
| Sampling Rate | 16khz |
| Encoding | Little Endian |

After the pre-processing, the system forwards the speech data to the Sphinx-4 recognizer. Though the frontend and decoder modules are pre-programmed by Sphinx-4, the knowledge base module (acoustic model, dictionary, and language model) are created by the user for their data. The model is programmed to access the following files: the *audio files*, *file_ids* list, the *transcripts*, *phone_list*, and the phonetic *dictionary*, that are created by the user in the transcription process, as shown in Section 3.5.

### 3.5    Speech-to-Text (STT) Transcription process

We adopted the following stages in processing speech-to-text from raw Kiswahili speech files, building an end-to-end system using guidance from other works (Dresing et al., 2015).

**Step 1** - Record the audio – in our case we used recordings in the existing Kencorpus project corpus (Wanjawa et al., 2022).

**Step 2** - Process the recordings, which we did using existing procedures adopted in other research (Moore et al., 2017). This is done by organizing data into a corpus to indicate where the pairs of audio and text files are stored, deciding on what to transcribe and saving the final voice data into a WAV file, that conforms to the specifications on Table 3.1 above.

**Step 3** - Create a text file called *file_ids*. In this created file, list the file_ids each on a separate line and then indicate its parent team folder, example:
*Speaker_1/0430_segment_1.wav*
*Speaker_1/0430_segment_2.wav*

**Step 4** - Create another text file called *transcript*. Start transcription by first making a rough transcription in which details such as prosody and pauses are not included. In this created file, transcribe each audio file in the order as per the *file_ids* of (3) above, for example:
*<s> words of segment 1 </s> (0430_segment_1)*
*<s> words of segment 2 </s> (0430_segment_2)*

**Step 5** - Create a text file called *dictionary*. In this file, all the words in the transcript with their corresponding pronunciations are listed. All duplicates are deleted, leaving only single instances of words. Phonetic pronunciation of a word is indicated against each word. The phonetic pronunciation is Capitalized. In our case we adopted a standard Kiswahili phoneme representation scheme. Augment the dictionary, we also added a list of fillers as part of the dictionary. This is the list of pauses/silence parts.

**Step 6** – Create a text file called *phone_list*. In this created file, all the phones available in the dictionary are listed without duplicates.

**Step 7** – Create a **language model** using CMU Sphinx toolkit (SRILM – SRI Language modelling toolkit) and associated coding and expose it to the files created in steps (1 to 6). This enables us to obtain the readout of performance metrics from the model.

**Sample of Kiswahili Transcription process**
Shown below is an example of the step-by-step process of transcription as described in Section 3.5 above, from raw speech files to final transcripts using the document processing and then the CMU Sphinx toolkit:

| Step 1 | use existing sound file existing on the KenCorpus dataset (Wanjawa et al., 2022) such as filename 0430_swa.MP4 |
|---|---|
| Step 2 | Process the audio file by deciding on the length of sound to use (which is not the entire content of the file). Save the file in WAV format e.g. 0430_swa_segment1.wav |
| Step 3 | Create the *file_ids* text file:<br>*Speaker_1/0430_segment_1* |
| Step 4 | Create the *transcript* text file for the identified section of the 0430_swa_segment1.wav audio file such as the sample shown below:<br>*<s> juu ya kitanda alilala mgonjwa ambaye kwa miezi saba sasa hajapata ashekali alizingirwa na walokole wakimuasaa Mterehemezi tena anakaribia kuwa mauti tofauti yake na maiti ilikuwa moja tu, uhai, sura yake iliyokuwa nzuri miaka ayami iliyopita sasa ilikuwa inatisha </s> (0430_swa_segment1)* |
| Step 5 | Create a *dictionary* for NLP using the Kiswahili phoneme representation (from Table 3.1), resulting into the following partial dictionary for NLP for Kiswahili language:<br>*juu         JH UU*<br>*ya          Y AH*<br>*kitanda     K IH T AH ND AH*<br>*alilala     AH L IH L AH L AH*<br>*mgonjwa     M G OH NJ W AH*<br>*uguzwe      UH G UH Z W EH*<br>*nilijizuia   N IH L IH JH IH Z UH IH AH* |
| Step 6 | Create a *phone_list* – the document containing all the phones available in the transcript without duplicates (from Table 3.1) and any fillers, for instance:<br>*AA AH B CH D DH EE EH.* |
| Step 7 | Use CMU Sphinx toolkit (SRILM) to create a *language model* and use the documents from the above processes to train and test the STT model<br>Get the test results in terms of performance metrics (WER, SER) |

## 4.     Experiments

We did our experiments from a baseline as a starting point and progressively changed the settings to determine the performance of the ASR over various experimental settings. The experiments done are described below. All our experiments followed the processes shown on the flowchart of Fig. 4.1 below, with the different parameters being described against each experimental setting.

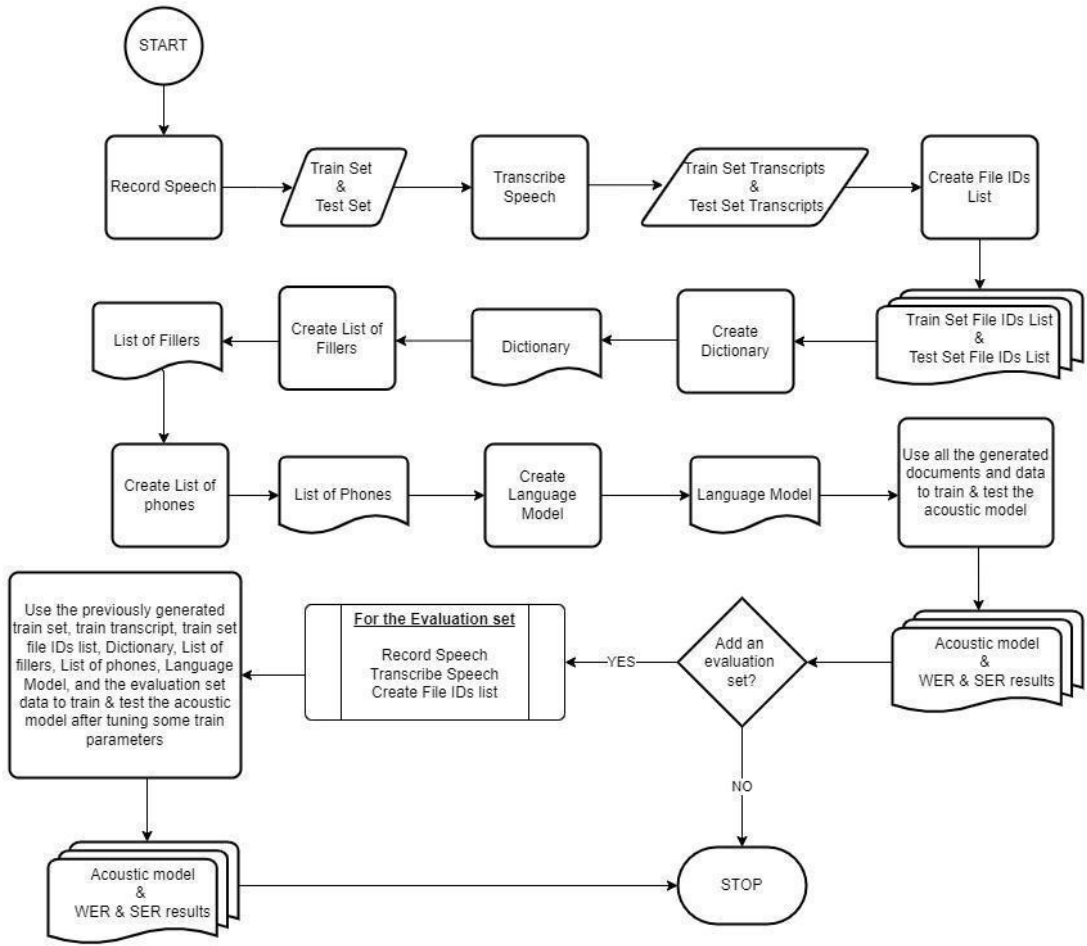

**Figure 4.1**: Flowchart for ASR of Kiswahili language (source: author)

The pseudocode for the STT showing the interaction of input files and CMU Sphinx toolkit is shown in Fig. 4.2 below.

1. ***Start***
2. Create audio files for ***training*** (.WAV)
3. Create audio files for ***TESTING*** (.WAV)
4. Process training audio file to create 4 other files: file_ids list, transcript, phone_list, phonetic dictionary (***training set files***)
5. Process testing audio file to create 4 other files: file_ids list, transcript, phone_list, phonetic dictionary (***TESTING set files***)
6. Use CMU Sphinx toolkit (SRILM) to create ***language model***
7. Use CMU Sphinx toolkit to train a ***training acoustic model*** based on the training language model and ***training set files***
8. Use CMU Sphinx toolkit to test the ***TESTING acoustic model*** based on the ***language model*** and ***TESTING set files***
9. Use CMU Sphinx toolkit to calculate the Word Error Rate (WER) and Sentence Error Rate (SER) of the ***TESTING set files***
10. IF there is evaluation set THEN create ***evaluation set files*** (similar to steps 5) else Stop

| 11 | *Use files in 10, together with those from 5 (**training + evaluation set files**) on CMU Sphinx to do train the new data files (step 7), then test the new data files (steps 8 and 9)* |
| 12 | ***Stop*** |

**Figure 4.2**: Pseudocode for ASR of Kiswahili language

### 4.1   Experiment 1 – 1 speaker, basic dictionary

The first experiment utilized a training set having one female speaker and 14min 50sec of speech data. The test set had 1 female speaker with utterances corresponding to 4min 20sec of speech data. The dictionary used for this experiment was the basic dictionary described in Section 3.3. The settings on the acoustic model were 8 Gaussian Mixtures and 2,000 senones. There was no evaluation set. The testing and evaluation metrics used were word error rate (WER) and sentence error rate (SER).

### 4.2   Experiment 2 – 1 speaker, ALFFA dictionary

In this experiment, the training and testing sets from Experiment 1 were maintained, with only the dictionary being changed. The dictionary used for this experiment was the ALFFA dictionary described in Section 3.3. We maintained all other settings of Experiment 1. The purpose of this experiment was to evaluate the effect of change of dictionary on the acoustic model performance.

### 4.3   Experiment 3 – 3 speakers, Augmented dictionary

This experiment used a training set consisting of 1 female and 2 male speakers and 1hr 22min 15sec of speech data. The testing set consisted of 1 male speaker with utterances equaling 20min 11sec of speech data. The dictionary used for this experiment was our new augmented dictionary described in Section 3.3. We maintained all other settings of Experiment 1. The purpose was to evaluate the effect of changing the dictionary to one that incorporates the missing phoneme representations evidenced in the ALFFA and basic dictionaries on the acoustic model performance.

### 4.4   Experiment 4 – 5 speakers, Augmented dictionary

This experiment used a training set of 4 males and 1 female with 5,337 speech files of 3hrs 33min 57sec as speech data. The testing set consisted of 2 male speakers with utterances equaling 1hr 15min 49sec of speech data. The dictionary used for this experiment was our new augmented dictionary. We maintained all other settings of Experiment 1. The purpose was to evaluate the effect of the addition of more speakers (voice variations) and audio hours on the acoustic model performance.

### 4.5   Experiment 5 – 26 speakers, Augmented dictionary

This experiment used the entire KenCorpus Kiswahili speech dataset of 26 speakers and 27hrs 31min 50sec of speech data. Out of this full dataset, our training set consisted of 10 females and 4 males with utterances corresponding to 17hr 30min 52sec of speech data. The testing set consisted of 6 females and 2 males with utterances corresponding to 9hrs 43min 46sec of speech data. The dictionary used for this experiment was the augmented dictionary. We maintained all other settings of Experiment 1. The purpose was to evaluate the effect of the addition of more speakers and audio hours on the acoustic model performance.

A new addition to this set, which is a key difference from the previous four experiments, was the introduction of an evaluation set that consisted of 3 females and 1 male corresponding to 17min 12 sec of speech data. The purpose was to evaluate the effect of the change in on the gaussian and senone parameters on the acoustic model performance. The full setup of the experimental parameters are shown on Table 4.1 below.

**Table 4.1**: Parameters set for Experiment 5

| Set | Female | Male | Duration |
|---|---|---|---|
| Training | 10 | 4 | 17:30:52 |
| Testing | 6 | 2 | 09:43:46 |
| Evaluation | 3 | 1 | 00:17:12 |
| Totals | 19 | 7 | 27:31:50 |

The first model for this experiment 5 (5a) had a training set with the baseline 8 Gaussian Mixtures and 2000 Senones for 10F/4M (17:30:52). The trained model was then tested on the test set that had 6F/2M (09:43:46). We then tested the effect of an evaluation set on the performance of our base system. We used the initial training set but changed the Gaussian Mixtures and Senones parameters as follows: 16 Gaussian Mixtures and 2500 senones. This trained model (5b) was then tested using the evaluation set 3F/1M (00:17:12), which turned out to provide the best performing model of all the experiments done.

## 5. Results and Discussions
### 5.1 Speech-To-Text (STT) Transcription model

The performance of the various models tested for Speech to text (STT) transcription as per the experiments done is shown on Table 5.1 below. The gender indicator provided for the speakers is either female (F) or male (M).

**Table 5.1** – Results of Kiswahili ASR model with different experimental settings

| Model | No. of speakers at Training | Training Time (H:M:S) | No. of speakers for Testing | Testing Time (H:M:S) | Acoustic model settings* | Has evaluation set | WER % | SER % |
|---|---|---|---|---|---|---|---|---|
| 1 | 1F | 0:14:50 | 1F | 0:04:20 | 8/2,000 | N | 100.0. | 100.0 |
| 2 | 1F | 0:14:50 | 1F | 0:04:20 | 8/2,000 | N | 93.00 | 100.0 |
| 3 | 1F/2M | 1:22:15 | 1M | 0:20:11 | 8/2,000 | N | 53.47 | 86.8 |
| 4 | 1F/4M | 3:33:57 | 2M | 1:15:49 | 8/2,000 | N | 30.21 | 47.7 |
| 5a | 10F/4M | 17:30:52 | 6F/2M | 9:43:46 | 8/2,000 | N | 20.10 | 52.3 |
| 5b | 10F/4M | 17:30:52 | 6F/2M | 9:43:46 | 16/2,500 | Y | 18.87 | 49.5 |

*(Gaussian Mixtures/senones)

**Experiment 1** – This experiment yielded an acoustic model with a WER of 100% and a SER of 100%. These low results were attributed to overfitting through the use of the same speaker for both the training and testing of the model. It was also observed that the dictionary used during the training was poorly crafted and led to limitations in the ability of the system to effectively understand the pronunciation of certain Kiswahili words spoken.

**Experiment 2** – This shift to a new dictionary format resulted in a WER of 93% and a SER of 100 %. While the WER and SER were still evidently high, we observed that a better dictionary could slightly improve the model's accuracy, despite the speaker being the same in both the training and testing stages.

**Experiment 3** – We observed significant improvement with WER of 53.47% and a SER of 86.8%. We attribute improvement over Experiment 1 and 2 due to improved variation of speakers through the addition of two male speakers for the training and also an unseen male voice for the testing as well as the use of the augmented dictionary that had a wider array of phonemes. These new phoneme representations ensured that most Kiswahili words and their pronunciations were captured, realizing better identification of Kiswahili words spoken.

**Experiment 4** – We noted significant improvement with an acoustic model performance with WER of 30.21% and a SER of 74.7%. While the acoustic model generated from this phase could effectively decipher a wide range of male voices when used within the ASR prototype, female voices evidenced poor transcription within the prototype. Additionally, it was evidenced that some of the audios were in wrong speaker folders which could have substantially lowered the chances of evidencing a WER of below 30%. CMU Sphinx requires audios to be correctly segregated according to speakers for ease of decoding the features of a particular speaker since voice features significantly vary between speakers. The performance improvement can also be attributed to the increase in audio hours as increased training data have been previously shown to improve machine learning models.

**Experiment 5** – Model 5 generated during this experiment was the best performing model. Unlike that of Experiment 4, this model could effectively decipher a wider range of both male and female voices. We observed WER and SER of 20.1% and 52.3% respectively in experiment 5a before incorporating the evaluation phase with the evaluation data. These results are attributed to the increased number of voices in the training set that ensured the generated model was not overfitting towards specific voices or their features and consequently pronunciation of Kiswahili words.

The settings of this final model of Experiment 5, incorporating the evaluation data (experiment 5b), was observed as the best combination of settings that could effectively accommodate a variety of speaker voices as input. This good performance is attributed to the higher number of senones that led to the model having a higher discrimination capability for spoken words. Nonetheless, we observe that when too many senones were used on a small corpus it would usually lead to the model being less generic to identify unseen speech, leading to a higher WER on unseen data.

A speech to text (STT) model with 16 Gaussian Mixtures and 2,500 senones, trained with both male and female voices over a training time of 27hrs 31min 50sec and exposed to an evaluation set of 17 minutes 12 seconds of both male and female voices, was observed as being the most optimal model that achieve the best performance where WER was 18.87% and SER was 49.5%.

### 5.2  Sample Transcription using the STT model of Experiment 5

Table 5.2 provides the sample results for the tests obtained when using our final speech-to-text (STT) model with the settings of experiment 5b. The test was done by a speaker not initially trained or tested during model creation and fine tuning. The decoded transcripts show highly accurate STT transcription. We note that the recognizer failed to transcribe some words within the speaker's speech, while other words were heard but wrongly decoded. These errors are shown as bold text on the 'transcript as per model' column.

**Table 5.2** – Actual and Decoded transcriptions from the developed STT model

| No. | Actual Transcript (as per the audio file) | Transcript as per the model |
|---|---|---|
| 1 | katika mwanamke bomba wiki hii tunamuangazia sharon chebet ambaye ni nahodha baharini chebet ana umri wa miaka ishirini na nane anasifika pakubwa kwa kazi yake | katika mwanamke bomba wiki hii tunamuangazia **sheria ubeti** ambaye ni nahodha baharini **bata** ana umri wa miaka ishirini na **nne anafika** pakubwa kwa kazi yake |
| 2 | wakulima wa mpunga eneo la nyando wanataka wadau kuwapa mbolea kukuza | wakulima **punga** eneo la nyando wanataka wadau kuwapa mbolea **kuzaa ndimu** |
|   | kilimo wanapania kuboresha kilimo chao kufuatia muingilio wa serikali | **wanapanya** kuboresha **ndimu** chao **kufuata** muingilio wa serikali |
| 3 | waziri wa utalii najib balala ahimiza jamii kuweka hatua madhubuti waziri balala haswa asisitizia kulindwa kwa simba humu nchini anasema idadi ya wanyama hawa imeendelea kupungua sana | waziri wa utalii **waji** balala ahimiza jamii kuweka **hata** madhubuti waziri balala haswa **kasisi** kulindwa kwa simba humu **chini** anasema idadi ya wanyama **uawa** imeendelea kupungua sana |
| 4 | mamlaka ya mazingira yatishia kufunga vichinjio kadhaa huko bungoma vichinjio hivyo vinadaiwa kukosa kuweka usafi wa kiwango cha juu | **miaka** ya mazingira yatishia kufunga vichinjio kadhaa huko bungoma vichinjio hivyo vinadaiwa kukosa kuweka usafi **mlango** juu |
| 5 | mwanamume yeyote yule mwenye pesa ana ugonjwa wa kutamani kila kiumbe ambacho ni cha jinsia tofauti naye yaani wanawake | mwanamume yeyote yule mwenye pesa ana ugonjwa wa kutamani kila kiumbe **choo** ni cha **jinsi** tofauti naye wanawake |
| 6 | wadau wataka elimu kuhusu athari za ngono ifunzwe shuleni wanafunzi wengi hawajui haki zao iwapo watadhulumiwa hafla ya kuhamasisha jamii yafanyika mjini kakamega | wadau wataka elimu kuhusu athari za ngono ifunzwe shuleni wanafunzi wengi hawajui haki zao iwapo **watahurumia** hafla ya kuhamasisha jamii yafanyika mjini kakamega |
| 7 | buriani bingwa wa sanaa msanii charles bukeko amezikwa kwao kaunti ya busia marehemu alijulikana na wengi kama papa shirandula atakumbukwa na wengi kwa mchango wake katika sanaa alikuwa mmoja wa wasanii maarufu wa runinga ya citizen | **biriani** bingwa wa **sana** msanii **chai mkeka** amezikwa kwao kaunti ya **mbuzi** marehemu alijulikana na wengi kama papa shirandula atakumbukwa na wengi kwa mchango wake katika **sana** alikuwa mmoja wa wasanii **arufu** wa runinga |
| 8 | mashaka ya linturi seneta aelekea mahakamani kuzuia kukamatwa na kushtakiwa hii ni kuhusiana na tuhuma za jaribio la ubakaji hotelini nanyuki seneta linturi atuhumiwa kujaribu kumbaka mwanamke huko | mashaka ya **vizuri** seneta aelekea mahakamani **kuvua** kukamatwa na **mshtakiwa** hii ni kuhusiana na tuhuma za **jaribu** la ubakaji hotelini nyuki seneta **vizuri hudumiwa jaribu** kumbaka mwanamke huko |
| 9 | chagua jibu ambalo si sahihi kulingana na aya ya pili madini huwa na faida chungu nzima katika bara la afrika pekee | **changaa** jibu ambalo **hii kuungana** na aya ya pili madini huwa na faida chungu **mzima** katika bara la afrika |
| 10 | mashujaa zaidi watuzwe katika hafla za kitaifa kichwa kinachofaa ufahamu huu ni siasa mbaya siku ya mashujaa siku ya madaraka lengo la mashujaa | **jaa** watuzwe katika **ghafla** za kitaifa kichwa **kinachovya kufahamu** huu ni sasa mbaya siku ya **shujaa** siku ya madaraka **shujaa** |

We note that existing automatic speech recognition (ASR) systems perform significantly better for high-resourced languages than for under-resourced languages. We observe significantly low WERs for English language of up to 5.1% (Kurata et al., 2017). English is assumed to be the most effectively handled language for ASR systems. On the other hand, under-resourced languages such as Somali, Hausa and Wolof have been observed to have WERs that are as low as 14.2%, 7.9% and 27.2% respectively, under different model training and testing conditions (Nimaan et al., 2006; Gauthier et al., 2016). Our WER of 18.87% for the low resource language of Kiswahili is therefore very good when compared to ASR performance of other low-resource languages. This is a good starting point and we believe that more training data in low-resource languages can improve such ASR performance even further.

The transcription process is very cumbersome, time consuming, and involves errors and deviations (Spencer et al., 2006; Dresing et al., 2015; McMullin, 2021). Our research also experienced the same challenges. One could take many hours, even a whole day, doing one transcription. This was due to the fact that some audio files were very long. Others were not very audible therefore the transcribers had to replay them many times when listening. It was also realized that many of the recorded audio files especially from the Coastal region of Kenya had the infusion of both the standard Kiswahili and coastal dialects that some transcribers used to the standard Kiswahili were not very familiar with. Some of the interviewed persons were also speaking very fast. In addition, in some instances, environments where recordings were taking place were somehow noisy. The long audio files had many pauses, and some had code switching and code-mixing influences.

## 6    Conclusion

This paper presents the phonemic representation and transcription of Kiswahili in the development of a speech corpus to train an ASR model for the language. The speech corpus contains 27hrs 31min 50sec of read out sentences and transcribed spontaneous speech. Besides, the study also discusses the addition of four new phonemes to the standard phonetic to accommodate the nuances in Kiswahili. The ASR model was trained using the phonetic set from Section 3.5. This yielded an initial WER and SER of 18.87% and 49.5%, respectively. Further research demonstrated that, in contrast to the initial training corpus, which was skewed toward male voices, advances could be made by integrating varied voices, especially female voices. The corpus's expansion to include female voices led to some improvement in model performance. Thus, for enhanced transcription, it is crucial to have a broad demographic in the voice corpus and complete phonemic representation of words within under-resourced languages.

The transcription process was not devoid of challenges including time-consuming transcription of lengthy speech files, some of which were inaudible due to noisy backgrounds. Besides, the infusion of both the standard Kiswahili and coastal dialects in the speech files, with some containing code switching, and fast speaking speakers exacerbated the transcription challenges. Strategies to circumvent some of these challenges included chunking the speech files into smaller lengths of one minute and below, using speakers located within their native language communities rather than in cosmopolitan areas, and managing the recording environments so that the resulting speech files were within permissible noise levels.

In general, this study highlights the numerous steps, peculiarities, and complexities in the transcription of Kiswahili in order to create a high-quality speech corpus for training an ASR model for any such under-resourced language.

**Acknowledgement**
This research was made possible by funding from Meridian Institute's Lacuna Fund under grant no. 0393-S-001 which is a funder collaboration between The Rockefeller Foundation, Google.org, and Canada's International Development Research Centre. We also acknowledge the inputs of the following research assistants for their contributions to the project: Japheth Owiny and Khalid Kitito.

# References

Akidah, M. A. (2013). Phonological and Semantic Change in Language Borrowing. The Case of Arabic Words Borrowed into Kiswahili. International Journal of Education and Research, 1 (4), 1-20. www.ijern.com

Ashton, E. O. (1951). Swahili Grammar. Kenya: Longman.

Baldi, S. (2012). Arabic Loans in East African Languages through Swahili: A Survey. FOLIA ORIENTALIA, 49, 37-52. http://journals.pan.pl/Content/85189/mainfile.pdf

Biswas, A., Menon, R., van der Westhuizen, E., & Niesler, T. (2019). Improved low-resource Somali speech recognition by semi-supervised acoustic and language model training. arXiv preprint arXiv:1907.03064.

Choge, S. (2009). Understanding Kiswahili Vowels. The Journal of Pan African Studies, 2 (8), 62-77. http://www.jpanafrican.org/docs/vol2no8/2.8_UnderstandingKiswahiliVowels.pdf.

CMUSphinx (2022). Open-source speech recognition toolkit. https://cmusphinx.github.io/

Deem, J. M., & Engel, S. A. (1988). Developing Literacy through Transcription. Journal of Basic Writing, 7 (2), 99-107. DOI:10.37514/JBW-J.1988.7.2.08.

Dresing, T., Pehl, T., & Schmieder, C. (2015). Manual (on) Transcription. Transcription Conventions, Software Guides and Practical Hints for Qualitative Researchers. 3rd English Edition. https://www.audiotranskription.de/wp-content/uploads/2020/11/manual-on transcription.pdf

Gangji, N., Pascoe, M., & Smouse, M. (2015). Swahili Research Development: Preliminary Normative Data from Typically Developing Pre-School Children in Tanzania. International Journal of Language and Communication Disorders, 50 (2), 151-164. DOI:10.1111/1460-6984.12118.

Gauthier, E., Besacier, L., & Voisin, S. (2016). Automatic speech recognition for African languages with vowel length contrast. Procedia Computer Science, 81, 136-143.

Gauthier, E., Besacier, L., Voisin, S., Melese, M., & Elingui, U. P. (2016b). Collecting resources in sub-Saharan African languages for automatic speech recognition: a case study of Wolof. In *10th Language Resources and Evaluation Conference (LREC 2016)*.

Gelas, H., Besacier, L., & Pellegrino, F. (2012). Developments of Swahili resources for an automatic speech recognition system. In *Spoken Language Technologies for Under-Resourced Languages*. http://www.ddl.cnrs.fr/fulltext/Gelas/Gelas_2012_SLTU.pdf

Getao, K., & Miriti, E. (2006). Creation of a Speech to Text System for Kiswahili. CMUdict. The Carnegie Mellon Pronouncing Dictionary. (2014). https://github.com/cmusphinx/cmudict

Habwe, J., & Karanja, P. (2004). Misingi ya Sarufi. Nairobi: Phoenix Press.

Higgins, H. A. (2012). Ikoma Vowel Harmony: Phonetics and Phonology. http://www.sil.org/systems/files/reapdata/12/83/13/128313646521624824655350 6614061697303369/e_book_43_Higgins_Ikoma_Vowel_Harmony.pdf.

International Rescue Committee (IRC) Research Tool Kit (2018). Qualitative Data Transcription and Translation. https://www.alnap.org/system/files/content/resource/files/main/ircqualitativetranscriptionand translationguidelines-ext.pdf

Kihore, Y. M., Massamba, D. P. B. na Msanjila, Y.P. (2009). Sarufi Maumbo ya Kiswahili Sanifu (SAMAKISA). Chuo Kikuu cha Dar es Salaam: TUKI.King'ei, K. (2010). Misingi ya isimujamii. Chuo Kikuu cha Dar es Salaam: Taasisi ya Taaluma za Kiswahili.

King'ei, K. (2010). Misingi ya isimujamii. Chuo Kikuu cha Dar es Salaam: Taasisi ya Taaluma za Kiswahili.

Kurata, G., Ramabhadran, B., Saon, G., & Sethy, A. (2017, December). Language modeling with highway LSTM. In 2017 IEEE Automatic Speech Recognition and Understanding Workshop (ASRU) (pp. 244-251). IEEE.

Massamba, David P. B. (2012). Misingi ya Fonolojia. TTK Dar es Salaam.

McMullin, C. (2021). Transcription and Qualitative Methods: Implications for Third Sector Research. International Society for Third-Sector Research. https://doi.org/1007/s11266-021-00400-3

Mgullu, R. S. (1999). Mtalaa wa Isimu. Nairobi: Longman.

Moore, E., & Llompart, J. (2017). Collecting, Transcribing, Analyzing and Presenting Plurilingual Interactional Data. In E. Moore & M. Dooly (Eds), Qualitative Approaches to Research on Plurilingual Education, 403-417. https://doi.org/10.14705/rpnet.2017.emmd2016.638

Mukiibi, J., Katumba, A., Nakatumba-Nabende, J., Hussein, A., & Meyer, J. (2022). The Makerere Radio Speech Corpus: A Luganda Radio Corpus for Automatic Speech Recognition. *arXiv preprint arXiv:2206.09790*.

Mwita, L. C. (2007). 'Prenasalization and the IPA'. UCLA Working Papers in Phonetics, 106, 58-67. http://www.ku.ac.ke/schools/humanities/images/stories/2016/scholarship.pdf.

Nchimbi, A. S. (1995). Je, Mang'ong'o ni Fonimu au si Fonimu katika Kiswahili Sanifu? Baragumu, 25-39. Maseno: Chuo Kikuu cha Maseno.

Nimaan, A., Nocera, P., & Bonastre, J. F. (2006). Automatic transcription of Somali language. In Ninth International Conference on Spoken Language Processing.

Obuchi, S. M., & Mukhwana, A. (2010). Muundo wa Kiswahili. Ngazi na Vipengele. Nairobi: A~Frame Publishers.

Okal, B. O. (2015). Uhakiki wa Fonimu na Miundo ya Silabi za Kiswahili. Kioo cha Lugha, 13, 103-124, (Print& Online), University of Dar es Salaam: Institute of Kiswahili Studies. https://www.ajol.info>kcl>article>view

Pantazoglou, F. K., Kladis, G. P., & Papadakis, N. K. (2018). A Greek voice recognition interface for ROV applications, using machine learning technologies and the CMU Sphinx platform. WSEAS Transactions on Systems and Control, 13, 550-560. Retrieved from https://www.researchgate.net/publication/331860083_A_Greek_voice_recognition_interface_for_ROV_applications_using_machine_learning_technologies_and_the_CMU_Sphinx_platform

Polome, E. C. (1967). Swahili Language Handbook. Washington: Center for Applied Linguistics.

Reddy, M. R., Laxminarayana, P., Ramana, A. V., Markandeya, J. L., Bhaskar, J. I., Harish, B., ... & Sumalatha, E. (2015, August). Transcription of Telugu TV news using ASR. In 2015 International Conference on Advances in Computing, Communications and Informatics (ICACCI) (pp. 1542-1545). IEEE. Retrieved from https://www.researchgate.net/publication/308872601_Transcription_of_Telugu_TV_news_using_ASR

Rincon, L. (2018). Guide for Transcribing Audio Records. DOI:10.13140/RG.2.2.30403.66086/1.

Singler, J. V. (2011). Pidgin and creoles in Anglophone West Africa. Maryland: University of Maryland Center for Advanced Study of Language, College Park. http://www.casl.umd.edu/sites/default/files/6_workshop_slides_singler.pdf

Spencer, M., & Howe, C. J. (2006). Optimal Strategies for Accurate Transcription. Literacy and Linguistics Computing, 21 (3). Doi:10.1093/llc/fqio30.

Trask, R. L. (1996). A Dictionary of Phonetics and Phonology. USA: Routledge.

Treece, R. (1990). Underspecifying Swahili Phonemes. https://kuscholarworks.ku.edu/bitstream/handle/1808/22932/MALC_1990_383-395_Treece.pdf?sequence=1

Trubetzkoy, N. S. (1969). Principles of phonology.

Verma, S. K. and Krishnaswamy, N. K. (1989). Modern Linguistics: An Introduction. OUP. Delhi.

wa Mberia, K. (2015). Misrepresentations and Omissions in Kiswahili Phonology. International Journal of Linguistics and Communication, 3 (1), 111. DOI:10.15640/ijlc.v3n1a12.

Wanjawa, B., Wanzare, L., Indede, F., McOnyango, O., Ombui, E., & Muchemi, L. (2022). Kencorpus: A Kenyan Language Corpus of Swahili, Dholuo and Luhya for Natural Language Processing Tasks. arXiv preprint arXiv:2208.12081.

Webb, V., & Kembo, E. (Eds.) (2000). African Voices: An Introduction to Language and Linguistics of Africa. Southern Africa: Oxford.

Whiteley, W. H. (1956). Studies in Swahili Dialect 1: Kimtang'ata. East African Swahili Committee.

Yilmaz, E., Biswas, A., van der Westhuizen, E., de Wet, F., & Niesler, T. (2018a). Building a unified code-switching ASR system for South African languages. arXiv preprint arXiv:1807.10949.

Yilmaz, E., McLaren, M., van den Heuvel, H., & van Leeuwen, D. A. (2018b). Semi-supervised acoustic model training for speech with code-switching. Speech Communication, 105, 12-22.

# Appendix 1

**Table A1** – The 29 Kiswahili consonant phonemes and example words

| Phoneme | Swahili word |
|---|---|
| /b/ | = **b**ora |
| /ch (tʃ)/ | = **ch**akula |
| /d/ | = **d**ada |
| /dh (ð)/ | = **dh**ambi |
| /f/ | = **f**ua |
| /g/ | = **g**ari |
| /gh(ɣ)/ | = **gh**ala |
| /h/ | = **h**araka |
| /j (ɟ)/ | = **j**ana |
| /k/ | = **k**ata |
| /l/ | = **l**adha |
| /m/ | = **m**kate |
| /mb/ | = **mb**oga |
| /n/ | = **n**ina |
| /nd/ | = **nd**oo |
| /ng/ | = **ng**oma |
| /ng'(ŋ)/ | = **ng'**ombe |
| /nj/ | = vu**nj**a |
| /ny (ɲ)/ | = **ny**a**ny**a |
| /p/ | = **p**aka |
| /r/ | = **r**amba |
| /s/ | = **s**asa |
| /sh(ʃ) | = **sh**amba |
| /t/ | = **t**ena |
| /th(θ)/ | = **th**amani |
| /v/ | = **v**aa |
| /w/ | = **w**eka |
| /y (j)/ | = **y**ai |
| /z/ | = **z**aa |

# Appendix 2

**Table A2** – Selected List of Kiswahili phonemes used in ASR

| Kiswahili | Phoneme Presentation NLP Word |
|---|---|
| alimuona | a l i m u o n a |
| alimwahidi | a l i m w a h i d i |
| alingara | a l i GG a r a |
| alinggara | a l i NN a r a |
| aliniambia | a l i n i a BB i a |
| aliniandikia | a l i n i a DD i k i a |
| Alionyesha | a l i o n y e SS a |
| aliowaacha | a l i o w a a CC a |
| alipoanzisha | a l i p o a ZZ i SS a |
| Zorah | z o r a h |
| Zoya | z o y a |